\documentclass[times, twoside]{zHenriquesLab-StyleBioRxiv}
\usepackage[colorlinks=true,
            allcolors=blue]{hyperref}
\usepackage{float}

\leadauthor{Abu Hweij}

\begin{document}

\title{Noninvasive Acute Compartment Syndrome Diagnosis Using Random Forest Machine Learning}
\shorttitle{ACS Machine Learning Diagnostic}

% Use letters for affiliations, numbers to show equal authorship (if applicable) and to indicate the corresponding author
\author[1*]{Zaina Abu Hweij\thanks{*Equal contribution}}
\author[2*]{Florence Liang}
\author[1*]{Sophie Zhang}

\affil[1]{Math Science Technology Program, Camas High School, Camas, WA 98607}\affil[2]{Irvine Valley College,  Irvine, CA 92618}

\maketitle

\begin{abstract}
Acute compartment syndrome (ACS) is an orthopedic emergency, caused by elevated pressure within a muscle compartment, that leads to permanent tissue damage and eventually death. Diagnosis of ACS relies heavily on patient-reported symptoms, a method that is clinically unreliable and often supplemented with invasive intracompartmental pressure measurements that can malfunction in motion settings. This study proposes an objective and noninvasive diagnostic for ACS. The device detects ACS through a random forest machine learning model that uses surrogate pressure readings from force-sensitive resistors (FSRs) placed on the skin. To validate the diagnostic, a data set containing FSR measurements and the corresponding simulated intracompartmental pressure was created for motion and motionless scenarios. The diagnostic achieved up to 98\% accuracy. The device excelled in key performance metrics, including sensitivity and specificity, with a statistically insignificant performance difference in motion present cases. Manufactured for 73 USD, our device may be a cost-effective solution. These results demonstrate the potential of noninvasive ACS diagnostics to meet clinical accuracy standards in real world settings.
\end {abstract}


\begin{corrauthor}
%\texttt{r.henriques{@}ucl.ac.uk}
fliang3\at ivc.edu
\end{corrauthor}

\section*{Introduction}
Acute Compartment Syndrome (ACS) is an orthopedic emergency that occurs when tissue pressure within a muscle compartment exceeds perfusion pressure. ACS typically necessitates immediate surgical intervention in the form of a fasciotomy to prevent irreversible nerve and muscle damage. In the United States, ACS affects approximately 26,500 individuals annually, with approximately 70\% of cases attributed to fractures \cite{giavia2015acute}. Despite its prevalence, ACS remains a challenging and complex condition to diagnose, even among expert surgeons.

The primary diagnostic method, known as the “5 P’s assessment” is subjective and clinically unreliable \cite{guo2019acute}. In severe trauma, patients are unable to differentiate the pain between fractures and increased compartmental pressure. This method may lead to misdiagnosis and unnecessary invasive compartment measurements. Moreover, the traditional invasive needling pressure measurement method is single-point, associated with discomfort, and not applicable in nonclinical settings \cite{li2019novel}.

As an alternative to invasive measurements, noninvasive ACS detection devices must be investigated. A solution that employs skin-to-surface pressure readings to detect surrogates of increased intracompartmental pressure may offer patients continuous ACS monitoring, seamless connectivity if multiple casts are worn, and straightforward usability. This study proposes a noninvasive skin-to-surface ACS diagnostic that employs a novel machine learning approach to provide patients with accurate diagnostics.

\section*{Related Work}
\subsection*{Acute Compartment Syndrome}
Acute Compartment Syndrome (ACS) is a condition characterized by increased tissue pressure within a closed muscle compartment, leading to muscle and nerve damage as well as impaired blood flow circulation \cite{pennmedicine2023}. This issue commonly occurs in orthopedic casts, such as the site of a fracture, where the injured limb is immobilized to aid the healing process. Following fractures, other common causes of ACS include burns, high energy trauma (crash injuries), drug overdoses, improperly applied casts, intense athletic activity, and improper positioning during surgery \cite{torlincasi2023acute}. 

The pathology of ACS is characterized by sharp increases in intracompartmental pressure or decreases in perfusion pressure, also known as delta pressure. Ideal intracompartmental pressure ranges from 0 to 8 mmHg. Pressures that indicate possible ACS begin from around 10 mmHg while the pressures for definite ACS requiring fasciotomy start from 40 mmHg and beyond. Clinicians generally adhere to the rule that a perfusion pressure, compartmental pressure minus diastolic pressure, of 30 mmHg or more is a distinct sign of ACS \cite{klenerman2007evolution}. To prevent and identify ACS, it is critical to be able to continuously monitor intracompartmental pressure or perfusion pressure in suspected cases.

Furthermore, studies have shown that external skin-to-surface pressure reading may be associated with intracompartmental pressure. Literature conducted with a pig skin model with imitated ACS has suggested that external and internal pressure is positively correlated \cite{gu2021new}. A similar study has also confirmed that the pressure between the skin and a cast can monitor intracompartmental pressure, with a correlation coefficient of 0.995 (P=0.000) \cite{uslu2000can}.

\subsection*{Invasive Diagnostics}
Measuring a patient’s compartmental pressure at the site of pain or muscle tensions is key in diagnosing ACS. A clinician does this in order to determine whether a fasciotomy is necessary to prevent permanent tissue damage. Often, they may follow the 5 P's assessment:  pain, pallor, paresthesias, paralysis, and pulselessness \cite{andrews1990neurovascular}. Patients who may be suspected of developing ACS may exhibit painful tension in muscles, a persistent ‘burning sensation,’ and paresthesias in addition to a recent orthopedic injury. However, there are certain limitations to this form of epidemiology. Children, critically ill patients, and those emerging from recent anesthesia may have a difficult time expressing these symptoms \cite{hammerberg2023}. All of these signs are hopelessly vague and there is little data to support the accuracy of these factors. At this stage, the patient’s clinician should take the patient’s compartmental pressure to confirm a diagnosis. There are two methods to measure compartment pressure that are frequently used and regarded as the “golden standard”: a syringe-based manometer (e.g. a Stryker device) and the wick or slit catheter technique \cite{allen2020classification}. 

The handheld manometer, particularly Stryker devices, have become a widely accepted and commercialized device because they are portable, simple, and relatively accurate. Such manometers consist of a needle to detect changes in compartmental pressure. They function by injecting a saline solution into a compartment and measuring the change in blood pressure in mmHg \cite{ULIASZ2003143}. Syringe-based manometers are used for independent measurements when a clinician needs to know the compartment’s pressure at a singular point in time. On the other hand, measurement methods using a catheter may be left within the patient for continuous monitoring. The slit catheter method involves inserting a fluid-filled tube into the compartment \cite{ULIASZ2003143}. The tube is connected to an arterial transducer which derives the compartment’s blood pressure by converting the analog pressure measurement into a readable voltage signal. 

Both devices are widely acclaimed for diagnosing ACS with relative accuracy, yet, there are still some limitations which should be improved upon. Each device should be calibrated and sterilized prior to each use. Multiple measurements of each site are suggested to ensure the diagnosis is accurate. Additionally, although there is little training required to use each device, technicians must learn the proper placement and usage of the probe to prevent misleading measurements. While injecting the saline solution into the site of possible ACS, technicians must make sure not to plunge too much liquid in too fast. Especially in a transducer-based device, a surplus of the saline solution could cause abnormal contractions as well as lead to inaccurate pressure readings and worsen ACS \cite{witthauer2020portable}. Furthermore, measurements taken from areas of abnormally high pressure  (< 5 cm) may result in a false positive \cite{hammerberg2023}. The wick catheter and slit catheter focus on increasing the surface area of pressure assessments to increase accuracy measures, however precision may still be inhibited by blood clots and air bubbles. Overall, further research is needed to better understand the clinical symptoms of ACS and improve the use-ability and accuracy of ACS diagnosing devices. 

\subsection*{Noninvasive Diagnostics}
While needle-based compartment pressure measurement is considered the gold standard for assessing ACS, noninvasive diagnostic tools have gained traction due to their potential for continuous monitoring, application in nonclinical settings, and early detection. Noninvasive approaches identify ACS through either the detection of decreased perfusion pressure or increased intracompartmental pressure \cite{Sellei21}. 

Infrared imaging and near-infrared spectroscopy (NIRS) present a promising solution towards measuring surrogates of perfusion pressure. In a clinical study, infrared imaging was shown to detect ACS by comparing the surface temperature of the proximal and distal leg. However, the application of this device may be limited due to the challenge of establishing precise temperature variance thresholds, especially in trauma patients \cite{katz2008infrared}. Additionally, NIRS-based devices, which aim to measure tissue oxygenation and perfusion pressure, have not demonstrated sufficient reliability for continuous monitoring because NIRS uses reflected light, which may be easily confounded by factors like skin abrasions, skin degloving, and muscle tissue hemorrhage \cite{schmidt2018continuous}. 

Current noninvasive approaches also measure surrogates of increased intracompartmental pressure. Skin-to-surface pressure readings, and tissue hardness measurement are emerging diagnostic tools. Although, literature considering systems employing handheld tissue hardness measurement note that present technology must be adapted to be more sensitive to small changes in pressure between the critical ranges of 20 and 50 mmHg \cite{joseph2006measurement}. Nevertheless, prior research has also investigated the use of a pressure sensor sleeve that can be worn underneath a cast \cite{ferrari2019sensei}. The skin-to-surface pressure reading system is capable of measuring the specified pressure range and interacts with an Android application to provide real time readings of when a sensor exceeds the 30 mmHg threshold. However, this device leverages a primitive detection algorithm that does not weigh the readings across the eight sensors used in the sleeve. This may lead to false positive pressure detections and ultimately unnecessary visits to physician offices and hospital emergency rooms. 

Recent research indicates that there is a current lack of reliable noninvasive diagnostic tools that can accurately pinpoint readings of individual sensors and ensure system connectedness if multiple casts are worn.  Patients may require monitoring in multiple limbs, and current technology is often fallible to ensure seamless connectivity and data aggregation from multiple devices. Among the various approaches, skin-to-surface pressure measurements show considerable promise to address past limitations. 
\subsection*{Flexible Sensor Technology}
Flexible pressure sensor technology holds high potential in noninvasive diagnostic tools for ACS.  In prior studies, researchers developed a flexible pressure sensor for measuring skin surface pressure in a pig-skin model with compartment syndrome \cite{gu2021new}. The sensor had a response area of 4×4 $mm^2$ and was encapsulated with PDMS to provide water resistance and biocompatibility. However, this limited sensing area may not be sufficient to capture pressure changes across larger regions or accurately represent the overall pressure distribution of a fracture site. Numerous sensors or sensors with wider response areas are necessary to achieve higher precision in ACS diagnostics.

Furthermore, the response units were prepared on a flexible foundation using a micro-nano structure direct writing technique with silver paste. In the study, the touch spot of the flexible pressure sensor was attached to the skin using medical tape. This implementation may be problematic in regards to secure adhesion and could potentially affect the accuracy of pressure measurements especially in longer duration uses. 

Literature has also considered the use of commercial Force Sensing Resistors \cite{ferrari2019sensei}. This system was reported to perform at a maximum deviation under 10\% over the 20-40 mmHg range. Testing also achieved a 91\% accuracy rate with ground truth negative readings at under 25 mmHg. However, this detection system held a 56.2\% specificity. Further work is necessary to determine how to reduce false positive rates in this design to prevent unnecessary clinical visits. We hypothesize that switching from a linear average algorithm as presented from this study to a machine learning algorithm to classify ACS detection based on sensor readings would greatly improve performance in these metrics.

The remainder of this paper is arranged as follows. Section 3 explains the methodology and validation.  Section 4 offers a discussion on the data. Finally, Section 5 presents conclusions to the data and offers final reflections for future research.
\section*{Materials and Methods}
\begin{figure}[H] % picture
    \centering
    \includegraphics[width=0.9\linewidth]{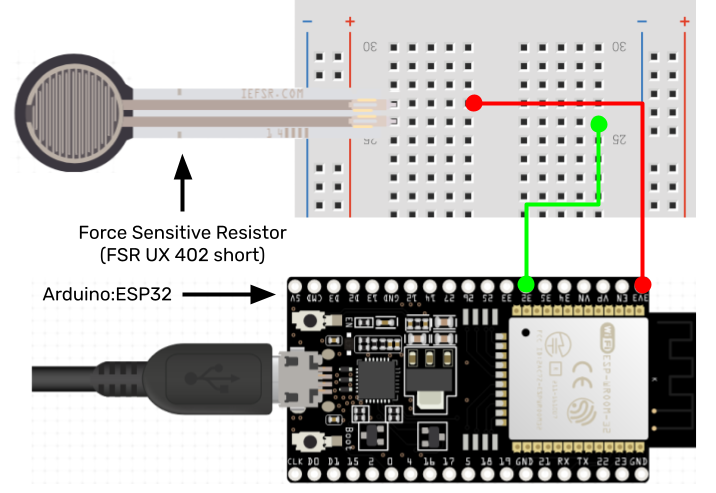}
    \caption{Visual schematic of fundamental electronic components. The force sensitive resistor (FSR) serves as the main way pressure data is collected while the Arduino:micro analyzes the data. The 10k Ohm resistor is used to reduce the amount of power fed into the FSR.}
    \label{fig:hardware}
\end{figure}

To improve system accuracy compared to prior noninvasive diagnostics, an ACS detection system that leverages machine learning was developed. The device was validated using a collected dataset containing ground truth ACS predictions and corresponding skin-to-surface pressure readings. A representation of the device illustrating the setup for just one force sensitive resistor (FSR) sensor is shown in Fig. \ref{fig:hardware}.

\subsection*{Hardware} 
To provide a comprehensive reading of the arm’s skin-to-surface pressure, five Interlink Electronics FSRs were sewn into a polyester biocompatible sleeve. Each sensor was placed to be in the approximate center of the anterior, lateral, superficial posterior and deep posterior compartments of a simulated lower leg. The sensors were wired to an ESP32 Arduino microcontroller, with built-in Bluetooth capability, to enable real-time data transmission and remote monitoring. The microcontroller was placed externally from the sleeve, maximizing patient comfort and preventing device breakage. The device was powered using a USB host device.

\subsection*{Dataset Collection}
\begin{figure}[H] % picture
    \centering
    \includegraphics[width=0.85\linewidth]{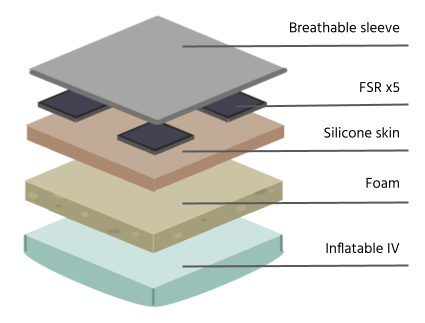}
    \caption{Cross-section of testing set-up. Acute compartment syndrome (ACS) was simulated artificially using an intravenous (IV) bag, foam and silicone skin. Ground truth pressure is the inflated pressure of the IV bag.}
    \label{fig:data_collection}
\end{figure}

As there is a lack of public data sets available on the diagnosis of ACS in relation to skin-to-surface pressure measurements, data points on the raw readings of five FSRs and the true prediction of ACS were gathered for motion and motionless scenarios. This dataset was collected by placing the diagnostic device on an inflated IV bag to simulate intracompartmental pressure. The IV bag was lined with a 21 mm polyethylene foam layer to simulate subcutaneous fat in the leg and a 2 mm silicone artificial skin. As shown in Fig. \ref{fig:data_collection}, the FSRs, sewn in the sleeve, directly touch the artificial skin. The IV bag was inflated from zero to 50 mmHg inclusive, by increments of 10 mmHg, and the corresponding FSR voltage readings were logged from the Arduino. A total of 400 data points with 80 data points for each pressure were recorded. This process was repeated to create another 400 point dataset containing recorded FSR measurements and ground truth pressure in a motion present scenario. Motion was simulated by manually raising and lowering the IV bag, by a 1 m vertical displacement, continuously throughout data collection. This was done to simulate real-life clinical settings such as moving the patient limb, transport conditions, and standing up. Ground truth pressures greater than or equal to 30 mmHg were classified as a positive case. Pressures below 30 mmHg were recorded as a negative ACS case. 

\subsection*{Random Forest Machine Learning}
\begin{figure}[H] % picture
    \centering
    \includegraphics[width=0.95\linewidth]{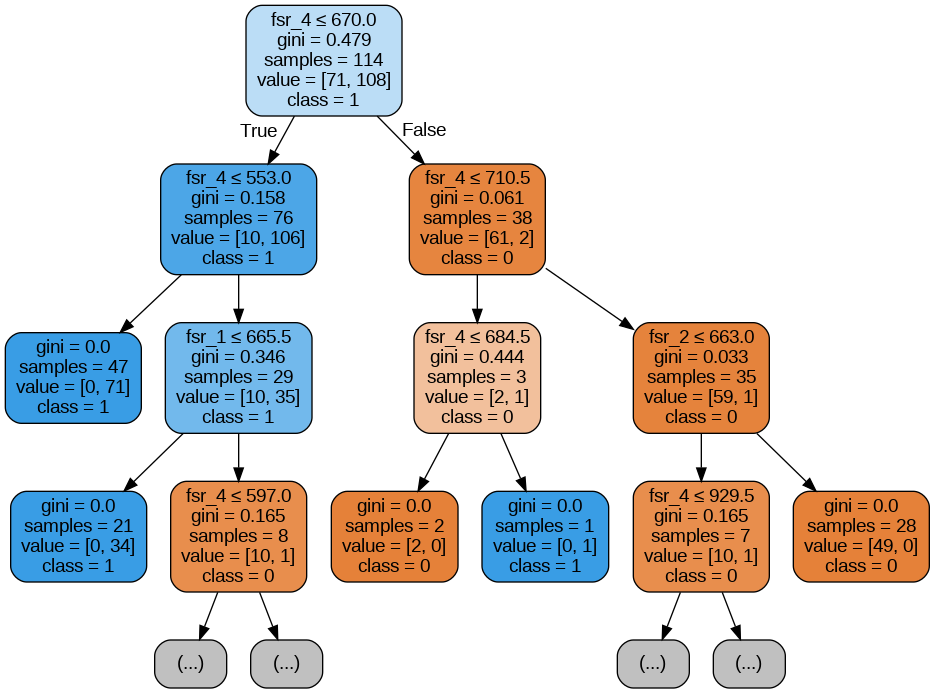}
    \caption{Single decision tree visualization. Darker shading indicates higher probability for the corresponding class, where blue is a positive detection and red is a negative detection.}
    \label{fig:decisionTree}
\end{figure}

To classify the detection of ACS using the FSR sensor measurements, a random forest classifier (RFC) machine learning model was deployed to the diagnostic. The model was trained and developed using the python library Scikit-Learn. The RFC algorithm was selected due to its recognized high accuracy in data science, and its ability to overcome noise in data \cite{reis2018probabilistic}. The model was trained using the ideal motionless data set. The test to train split of the data was 20\% to 80\%. The model was then evaluated using 20\% of the motion data set to assess generalization of the model in the alternate use case. The input parameters were the raw voltage readings of five FSR sensors and the output was a binary classification of ACS. No feature preprocessing or scaling was performed on the data-set. The RFC was constructed using the standard number of 100 decision trees and all hyperparameters were left at default values. The max depth of a decision tree was abbreviated to be five to prevent model overfitting of data. This was also done to improve model generalization for a motion present case. An abbreviated single decision tree from the forest is visualized in Fig. \ref{fig:decisionTree}, where darker shaded boxes indicate a higher probability for the detection class.

\begin{figure}[H] % picture
    \centering
    \includegraphics[width=1\linewidth]{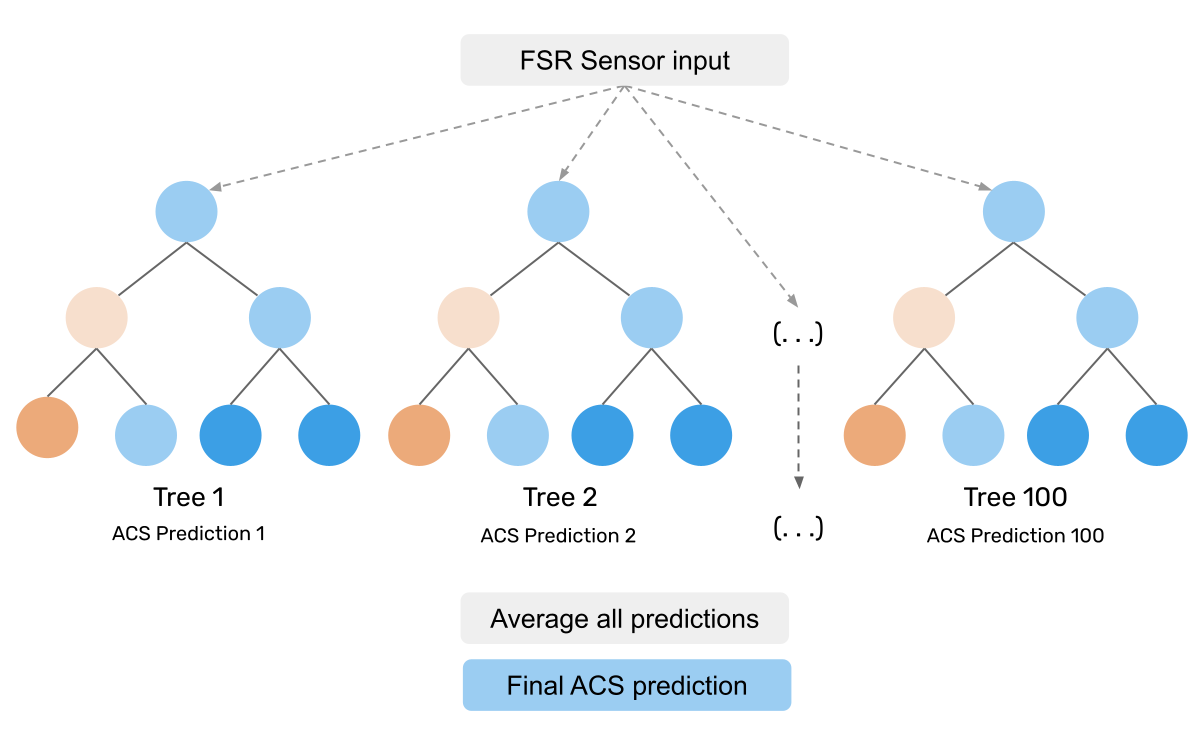}
    \caption{RFC algorithm final classification using decision trees. The model averages predictions across 100 decision trees.}
    \label{fig:treeDiagram}
\end{figure}

The gini index shown calculates the probability an instance is wrongly classified when it is randomly chosen.
As shown in Fig. \ref{fig:treeDiagram}, the RFC algorithm utilizes a collection of decision trees and then applies majority voting on the trees to assign a final classification for the given set of inputs.

The model’s utilization of numerous decision trees avoids potential issues with any overfitting. The model was then exported to the ESP32 Arduino. In real-time use cases, the Arduino read raw data from the FSRs, processed the pressure measurements using the RFC model, and exported the final diagnostic to a ReactJS web application via bluetooth. 

\subsection*{Diagnostic Evaluation}
Instances of true and false positive and negative predictions were recorded for the ACS machine learning diagnostic using the test portion of the prior collected dataset. The device’s performance in classifying the presence of ACS, in terms of accuracy, precision, sensitivity (also known as recall), specificity, and F1 score were evaluated. These metrics are detailed in equations (1) - (5), where TP denotes true positives, TN denotes true negatives, FP denotes false positives, and FN denotes false negatives.

\begin{equation}
\text{ACC} = \frac{\text{TP} + \text{TN}}{\text{TP} + \text{TN} + \text{FP} + \text{FN}}
\end{equation}

\begin{equation}
\text{PRE} = \frac{\text{TP}}{\text{TP} + \text{FP}}
\end{equation}

\begin{equation}
\text{SEN} = \frac{\text{TP}}{\text{TP} + \text{FN}}
\end{equation}

\begin{equation}
\text{SPE} = \frac{\text{TN}}{\text{TN} + \text{FP}}
\end{equation}

\begin{equation}
\text{F1} = \frac{2 \cdot \text{PRE} \cdot \text{SEN}}{\text{PRE} + \text{SEN}}
\end{equation}

\begin{figure*}[t] % picture
    \centering
    \includegraphics[width=0.8\linewidth]{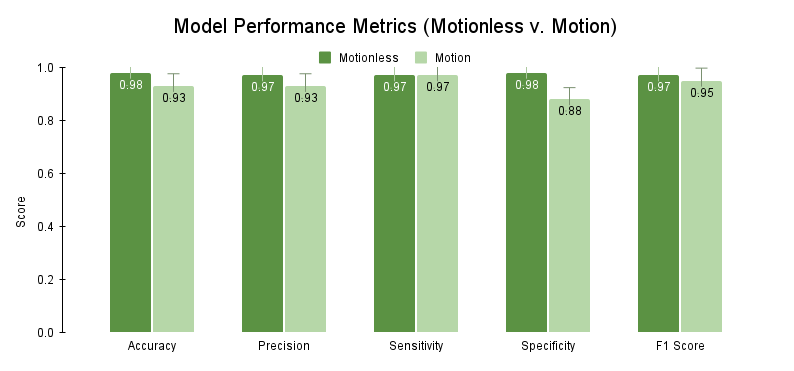}
    \caption{Comparison of model performance metrics in motion and no motion present scenarios. Higher scores indicate greater performance. Accuracy, precision, sensitivity, and specificity tended to be of stronger values from the range of 0 to 1 in both test cases. Performance during motionless diagnosis was found to be statistically insignificant compared to the motion present case. Ten percent error bars were used to determine significance.}
    \label{fig:performance}
\end{figure*}

\section*{Results and Discussion}
Our model’s performance in motionless and motion scenarios was assessed using key metrics, including accuracy, precision, sensitivity, specificity and F1 score. The accuracy and precision during the model’s evaluation in the motionless case was notably high at 0.98 and 0.97 respectively. Although accuracy and precision is higher in the motionless case, these metrics are not statistically significantly different from the motion case, when using ten percent error bars. The model in both cases achieved a sensitivity of 0.97, which indicates that fewer cases of ACS are missed. The model also demonstrated a high motionless specificity of 0.98 and an acceptable motion specificity of 0.88. Furthermore, the F1 score reached 0.97 and 0.95 in motionless and motion scenarios, which signals a well-balanced performance between precision and sensitivity. All these metrics are presented in Fig. \ref{fig:performance}, where accuracy, precision, sensitivity, specificity, and F1 score were consistently on the higher range from 0 to 1 in the two scenarios.

Out of all metrics, the model exhibited a lower specificity in the motion case. This suggests a possible challenge in correctly ruling out ACS. Nevertheless, in comparison to prior literature that assessed a diagnostic with similar hardware but a simple average algorithm among pressure readings, our study still proposed a 57\% increase in specificity \cite{ferrari2019sensei}. This is likely because machine learning is more capable of discerning complex relationships between the sensors compared to a simple linear approach. 

Notably, our device's accuracy is comparable to the industry standard arterial line manometer, both sharing the same accuracy score \cite{boody2005accuracy}. The high performance in accuracy indicates the potential of our system as a clinically reliable diagnostic tool.

\begin{figure}[H]
    \centering
    \includegraphics[width=1\linewidth]{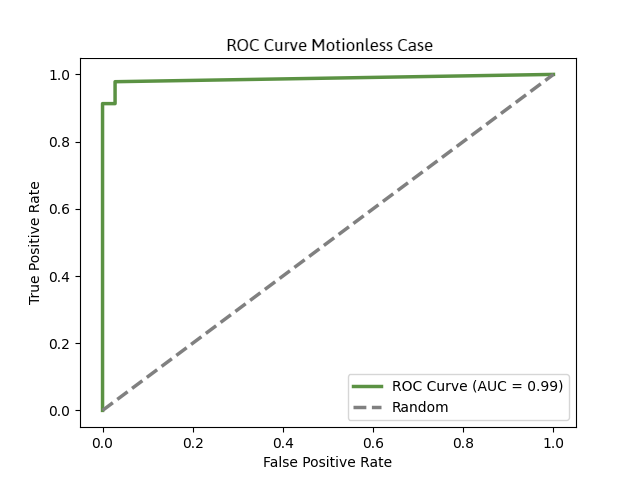}
    \caption{The generated receiver operating characteristic (ROC) curve of our detection model in the motionless case. The upper-left corner convex shape and related area under curve metric indicates strong performance in distinguishing between true positives and false positives.}
    \label{fig:ROC_curve_motionless}
\end{figure}

Moreover, in Fig. \ref{fig:ROC_curve_motionless}, a Receiver Operating Characteristic (ROC) curve was generated to assess the diagnostic performance of the model in the motionless scenario. The ROC curve visually represents the trade-off between sensitivity and specificity across different threshold settings in the model’s output probability for a binary classification of ACS. The convex shape towards the upper-left corner of our model’s ROC curve suggests a robust ability to correctly classify positive instances while minimizing false positives. 

\begin{figure}[H]
    \centering
    \includegraphics[width=1\linewidth]{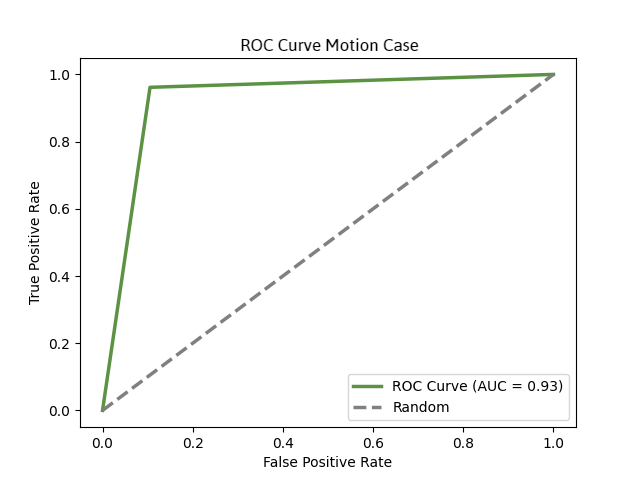}
    \caption{The generated receiver operating characteristic (ROC) curve of our detection model in the motion case. The upper-left corner convex shape demonstrates that the classifier’s performance is not due to random chance.}
    \label{fig:ROC_curve_motion}
\end{figure}

This trend is also shown for the ROC curve of the motion scenario shown in Fig. \ref{fig:ROC_curve_motion}. This convexity signifies a favorable model performance, as our model outperformed a random classifier, depicted by the dashed gray line, and approached the ideal scenario of perfect classification of the condition in both test cases. The area under curve (AUC) is greater in the motionless case compared to the motion present scenario. These findings are expected as movement may interfere with FSR measurement and introduce further noise in the data. 

Altogether, the total manufacturing cost to produce a user ready product was 73 USD. In reference, the industry standard Stryker intracompartmental pressure system is commercially available for 1575 USD \cite{ULIASZ2003143}. Furthermore, our device is 97\% cheaper than a 2019 study proposing a noninvasive diagnostic utilizing similar technology that held a total cost of  144.31 USD \cite{ferrari2019sensei}. Our device's cost-effectiveness is most likely due to the use of commercially available FSRs rather than self manufactured ones. 

In conclusion, our diagnostic model demonstrated exceptional performance metrics, including high accuracy, precision, sensitivity, and F1 score, with a slight trade-off in specificity in the motion case. The cost-effective manufacturing of our device makes it a compelling and economic alternative to the current invasive gold standard.

\section*{Conclusion}
In this study, we developed a noninvasive ACS diagnostic system that utilizes skin-to-surface pressure measurement and RFC machine learning for real-time and continuous detection. To address the lack of public data sets on ACS, our work contributed a dataset containing FSR readings and corresponding simulated intracompartmental pressure in both motionless and motion settings. Validation of our diagnostic demonstrated high performance in accuracy, precision, sensitivity, specificity, and F1 score. Notably, our testing suggests that our noninvasive device can achieve accuracy levels comparable to the clinical invasive gold standard and may be able to serve as a cost-effective alternative. Furthermore, our device was able to accurately predict ACS in motion present settings with an insignificant difference compared to motionless detection. This finding demonstrates that a noninvasive diagnostic can be applied to real-world scenarios where continuous pressure monitoring includes periods where the patient is being transported. Additionally, these results indicate that a machine learning noninvasive diagnostic can address past limitations of invasive devices becoming fallible in motion present scenarios \cite{MerleComeau}. Comprehensively, our study presents a promising noninvasive ACS diagnostic that holds the potential to meet clinical standards, become adopted into the standard of care and improve the patient experience.

\section*{Future Work}
As our device was tested using an artificial simulation of intracompartmental pressure, the generalizability of our results to real-world scenarios may be constrained. A continuation of our study could entail in vivo testing to thoroughly validate the efficacy of our diagnostic. Further work including animal models would aid to confirm the device’s accuracy. Additionally, extended tests analyzing the device’s performance in long-term continuous applications have the potential to provide greater insight into the device’s capabilities. Overall, future work is necessary to allow the translation of our device into medical applications.

\begin{acknowledgements}
This work was funded by the Camas High School Math Science and Technology Program.
\end{acknowledgements}

\section*{Bibliography}
\nocite{*} 
\bibliography{references}
\end{document}